\documentclass{llncs}
\usepackage{graphics}
\usepackage{amsmath}
\usepackage{epstopdf}
\usepackage{url}
\usepackage{amssymb}
\usepackage{graphicx}
\usepackage{color}

\usepackage{amssymb}
\usepackage{graphicx}

\usepackage{epsfig}
\usepackage[ruled,vlined]{algorithm2e}
\usepackage{amsmath}
\usepackage{subfigure}

\usepackage{enumitem}
\usepackage[hidelinks]{hyperref}

\usepackage[T1]{fontenc}
\usepackage[utf8]{inputenc}
\usepackage{array}
\usepackage{refstyle}
\usepackage{multirow}

\usepackage[author={Louis Chartrand}]{pdfcomment}

\begin{document}

\mainmatter

\title{Detecting Large Concept Extensions \\for Conceptual Analysis}

\author{Louis Chartrand\inst{1},
Jackie C.K. Cheung\inst{2},
Mohamed Bouguessa\inst{1}}

\institute{University of Quebec at Montreal\\
Department of Computer Science\\
Montreal, Quebec, Canada\\
\and
McGill University \\
School of Computer Science\\
Montreal, Quebec, Canada}

\toctitle{Detecting large concept extensions for conceptual analysis}
\tocauthor{Louis Chartrand,
Jackie C.K. Cheung,
Mohamed Bouguessa}

\maketitle

\begin{abstract} When performing a conceptual analysis of a concept,
philosophers are interested in all forms of expression of a concept in a
text---be it direct or indirect, explicit or implicit. In this paper, we
experiment with topic-based methods of automating the detection of concept
expressions in order to facilitate philosophical conceptual analysis. We propose
six methods based on LDA, and evaluate them on a new corpus of court decision
that we had annotated by experts and non-experts. Our results indicate that
these methods can yield important improvements over the keyword heuristic, which
is often used as a concept detection heuristic in many contexts.
While more work remains to be done, this indicates that detecting concepts
through topics can serve as a general-purpose method for at least some forms of
concept expression that are not captured using naive keyword approaches.
\end{abstract}


\section{Conceptual Analysis as a Computational Linguistics Problem\label{sec:Intro}}

Conceptual analysis in philosophy can refer, in a technical sense, to the
discovery of \emph{a priori} knowledge in the concepts we share
\cite{Chalmers2001_CHACAA}, \cite{Jackson1998_ACFMT},
\cite{Laurence2003_LAUCAC}. For instance, philosophers will say that ``male
sibling'' is a proper analysis of the concept BROTHER, because it decomposes its
meaning into two other concepts: a brother is nothing more and nothing less than
a male sibling.
Doing so allows us to make explicit knowledge that is \emph{a
priori}, or in other words, knowledge that is not empirical, that can be acquired without
observation: for instance, the knowledge that a brother is always a male.
In a
broader sense, it can refer to the philosophical methods we use to uncover the
meaning and use of a concept in order to clarify or improve it
\cite{NovaesForthcoming_NOVCEF}, \cite{Haslanger2012_HASRRS}. Given philosophy's
focus on conceptual clarity, the latter has been ubiquitous in practice. These
methods usually seek to make explicit key features of the concept under
scrutiny, in order to construct an account of it; be it a formalized
representation that can be expressed in terms of necessary and sufficient
conditions, or a more intuitive and pragmatic account of it.

Among the empirical sources upon which conceptual analysis relies, textual data
is one of the most important. While armchair philosophy (which relies on thought
experiments and intuitions) helps one give a better account of his or her own
concepts, contact with texts provides an essential perspective. As a result,
philosophers often build {\em corpora}, i.e. databases of texts that likely use
or express a particular concept that is undergoing analysis.

In philosophy as elsewhere, corpora need to be broad enough to cover all the
types of usages of the concept under scrutiny, lest the analysis fails to be
exhaustive. In other disciplines of social science and humanities, the necessity
of grounding analysis in corpora has lead researchers to harness text mining
and natural language processing to improve their interpretations of textual
data. Philosophy, however, has remained untouched by those developments, save
for a few projects \cite{Braddon-Mitchell2009-BRAITC-2}, \cite{meunier2005classification}.

One important obstacle to the adoption of those methods in philosophy lies in
the lack of proper concept models for conceptual analysis. Keyword approaches to
identifying concepts can run into ambiguity problems, like polysemy and
synonymy. Furthermore, they can only detect explicit concepts, whereas passages
where a concept is latent are bound to also interest the analyst. Latent concept
approaches, such as latent semantic analysis (LSA) \cite{deerwester1990indexing}
or latent dirichlet allocation (LDA) \cite{blei2003latent}, can work to
alleviate ambiguity problems and detect latent semantic expressions, but the
dimensions they generate (``concepts'' in LSA, ``topics'' in LDA) are thematic,
not conceptual. While concepts typically refer to abstract entities or entities
in the world, themes, topics and other thematic units are discursive: they can
only describe features and regularities in the text.

The problem we address in this paper is that of retrieving textual segments
which are relevant to philosophical conceptual analysis. Considering that
conceptual analysis is interested in the entire set of textual segments where a
queried concept is present in any form, the task at hand is to detect segments
whose discourse expresses, implicitly or explicitly, a queried concept. Our
concept detection problem distinguishes itself from traditional information
retrieval problems in that the aim is to retrieve text segments where the
queried concept is \emph{present}, rather than text segments that are
\emph{relevant} to the queried concept. In the context of a relevance search,
the inquirer will look for the minimum number of documents that can give the
maximum amount of generic information about the queried concept; for instance, a
web search for ``brother'' will likely return dictionary definitions and the
Wikipedia entry for this word. In the context of a presence query, the inquirer
will look for all of the documents where the queried concept is present, thus
enabling a more subtle understanding of the concept in all its shades. A search
for the presence of the concept BROTHER might thus return texts in genetics or
inheritance law as well as implicit evocations of brotherly love in a play. On
the other hand, this concept detection problem also differs from entity
recognition or traditional concept mining, as the concept does not need to be
associated with a word or an expression. While these problems focus on one
particular way a concept can be expressed, conceptual analysis will be
interested in any kind of expression of a concept, be it direct or indirect,
explicit or implicit.

As such, in \secref{ConceptDetection}, we clarify what counts as concept
expression for the sake of conceptual analysis, and we distinguish it from other
similar notions. In \secref{Method}, we describe methods to detect a queried
concept's expression in textual segments from a corpus. In \secref{Exp}, we
present how these methods were implemented and tested, including how an
annotated corpus was built, and in \secref{Results}, results are laid out.
Finally, in \secref{Discussion}, results are discussed, in a bid to shed
light on the underlying assumptions of the methods employed.

\section{Concept Detection\label{sec:ConceptDetection}}


While conceptual analysis can take many forms, it can always be enhanced by
taking empirical data into account. Philosophers who set out to make a concept's
meaning explicit through its analysis typically already possess the said
concept, and can thus rely on their own intuitions to inform their analysis.
However, their analysis can be improved, both in terms of quality and in
validity, by being compared with other sources. This explains, for instance, the
appeal of experimental philosophy, which has developed in the last 15 years as a
way of testing philosophical intuitions using the tools of cognitive and social
psychology\cite{Knobe2008xphi}. However, these inquiries have their limits: the intuitions they aim
to capture are restricted to a specific time and scope, as they are provoked in
an artificial setting. Textual corpora give us the opportunity to study
concepts in a more natural setting, and in broader populations, or in
populations which are hard to reach via conventional participant recruitment
schemes (experts, authors from past centuries, etc.).

In order to use data from textual corpora, philosophers now have access to the
methods and techniques of computer-assisted analysis of textual data. Those
methods and techniques are both numerous and diverse, but there are some common
characteristics. For instance, they typically involve various steps, which, together, form
treatment chains \cite{fayyad1996data}, \cite{meunier2005classification}: textual data
are preprocessed (cleaning, lemmatisation, etc.) and transformed into suitable
representations (e.g. vector-space model); then, specific treatment tasks are
performed, and finally their output is analysed and interpreted. Furthermore,
concepts must be identified in the text, in order to extract their associations
to other features that can be found in textual data, such as words, themes and
other concepts.


One way of identifying a concept in the text is to identify textual segments
in which it is expressed. This expression can take many forms: it can be a word
that explicitly refers to a concept in a very wide variety of contexts ("moose"
for MOOSE), a description ("massive North American deer"), or
embedded in an anaphoric reference ("the animal that crossed the street",
"{\em its} habitat"). It can also be expressed in such a way that it is not tied
to any specific linguistic expression. For instance, it can appear in the
background knowledge that is essential in understanding a sentence (for instance,
in talking about property damage that only a moose could have done), or in relation to the ontological hierarchy (for instance, the concept MOOSE can be expressed when talking about a particular individual moose, or when talking about cervidae).

Our objective here is to test methods of identifying such expressions in textual
segments. In other words, our goal is to detect, within a corpus, which
passages are susceptible to inform our understanding of how the concept is
expressed in a corpus. As such, concept detection can be seen as a useful step
in a wide variety of computer-assisted conceptual analysis methods. For
instance, it can act as a way of reducing the study corpus (i.e., the corpus on which a concepual analysis is based) to make it more
digestible to a human reader, or it can signal that the semantic content of the
segments where the queried concept is detected is likely to be related to the
concept, and thus enable new ways of representing it.

Because conceptual analysis is focused on a concept's expression
in discourse, concept detection is interested in its presence in discourse. This
can mean that the concept is explicitly present, and that it can be
matched to a word or an expression, but this presence can also
be found in other ways. It can be present in the postulates of the
argumentation, without which the passage would be impossible to understand.
(For instance, talk of incarceration takes on a very different meaning
if we lack the concept of sentencing for a crime). It can be a hypernym
to an explicit hyponym, if its properties, expressed to the hyponym,
are important enough to the discourse content that we can identify
the hypernym as a relevant contributor to the proposition. It can
be present in the theme that's being expanded in the passage. It can
be referenced using a metaphor or an anaphor. To synthesize, this
criterion can be proposed: {\em a concept is expressed in a textual segment
if and only if possession of a concept is necessary to understand the content
of the segment}.


Concept detection is similar to other popular problems and projects that have
been developed within NLP. However, important distinctions justify our treating
it as a different kind of problem.

For instance, concept detection differs from information retrieval (IR) in that presence, rather than relevance, is what we are looking for. For
instance, while IR might be interested in giving priority to text segments where
a queried concept is central, this is of little importance to a conceptual
analysis, as salient and less salient expressions of a concept are likely to
give different yet equally important dimensions of a concept of interest.
Conversely, while IR is interested in relevance of a document to a concept even
if it is absent, such a rating is meaningless if one is only looking for
presence or absence.

It also differs from other tasks which are geared towards presence detection,
such as named-entity recognition or coreference resolution. While a concept can
be present because a word or expression directly refers to it, it is not absent
because no such expression exists in a sentence or another textual unit. In
other words, a concept can be present in a text segment even if no single word
or expression refers to it. It can be present in virtue of being part of
the necessary background knowledge that is retrieved by the reader to make sense
of what she or he is reading. Concept detection, as we mean it, should detect both
direct and indirect presence of concepts.

\section{LDA Methods for Detecting Concepts\label{sec:Method}}

The presence of a concept as described in \secref{ConceptDetection} can
therefore be expressed in various ways: direct explicit reference, anaphorical
or metaphorical reference, implicit argumentative or narrative structures, etc.
In order to detect these different types of presence, one may expect that we
should fragment the task of concept detection into more specific tasks attuned
to specific types of presence. In other words, we could detect concept presence
by running various algorithms of named-entity recognition or extraction,
coreference resolution, topic models, etc. Each of these would detect a specific
way in which a concept can become present in a text, and we would rule
that a concept is present in a text segment if it has been detected with any of
the methods employed. However, not only is such an approach potentially very
time consuming, it makes it very hard to have a constant concept representation:
these various algorithms will accept different types of representations of the
queried concept, and as such, it will be hard to guarantee that they are all
looking for the same concept.

One way around this problem is to hypothesize that while these various
expressions of a concept are expressed in different ways, they may be
conditioned in similar ways by latent variables. We suppose, in this way, that
topics---i.e. underlying discursive and narrative constructions which structure
a text, cf. \cite{blei2003latent}---are such latent variables that condition the
expression of words and concepts alike. For instance, if the topic ``family
dinner'' is present in a text excerpt, it makes it likely for words such as
``table'', ``mother'', ``brother'' to be present, and unlikely for words such as
``clouds'' or ``mitochondria'' to be present; and in a similar fashion, concepts
such as FOOD, BROTHER and MOTHER are likely to be expressed and concepts such as
ORGANELLE and CLOUD are likely absent.

Therefore, given a concept expressed as a word that is typically associated
with it, we can find topics in which it is expressed, and use those
topics to find the textual segments where it is likely to be present.

We implement this approach using two different algorithms for learning an LDA
model, one that is based on Hoffman's online learning
algorithm \cite{hoffman2010online} and one that is based on Griffiths \&
Steyvers's Gibbs sampler \cite{griffiths2004finding}.

\subsection{Online Learning}

Hoffman's algorithm \cite{hoffman2010online} is an online variational Bayes
algorithm for the LDA. As such, it relies on the generative model
that was introduced by Blei \cite{blei2003latent}.

Blei's model uses this generative process, which assumes a corpus
$D$ of $M$ documents each of length $N_{i}$:
\begin{enumerate}
\item Choose $\theta_{i}\sim\mathrm{Dirichlet}\left(\alpha\right)$, where
$i\in\left\{ 1\ldots M\right\} $, the topic distribution for document
$i$
\item Choose $\phi_{k}\sim\mathrm{Dirichlet}\left(\beta\right)$, where
$k\in\left\{ 1\ldots K\right\} $, the word distribution for topic
$k$
\item For each of the word positions $i,j$, where $j\in\{1,\dots,N_{i}\}$
, and $i\in\{1,\dots,M\}$:

\begin{enumerate}
\item Choose a topic $z_{i,j}\sim\mathrm{Multinomial}(\theta_{i})$.
\item Choose a word $w_{i,j}\sim\mathrm{Multinomial}(\varphi_{z_{i,j}}).$
\end{enumerate}
\end{enumerate}

Here, $\alpha$ and $\beta$ are parameters of the Dirichlet prior on the
per-document topic distributions and on the per-topic word distribution
respectively; $\theta_i$ is the topic distribution for document $i$; and
$\phi_k$ is the word distribution for topic $k$.

Through online stochastic optimization, the online LDA algorithm learns $\theta$
(the topic distributions for each document) and $\phi$ (the word distribution
for each topic). Thus, it is possible to know which topics are likely to be
found in each document, and which words are likely to be found for each topic.

Using this information and given a queried concept represented as
a word, we can use $\phi$ to find the topics for which it is among the
most important words, relatively, and then use $\theta$ to find the
documents in which these topics have a non-negligible presence. We
thus have a set of documents which are likely to contain the queried
concept.

\subsection{Gibbs Sampling-LDA}

While it uses the same LDA model, Griffiths \& Steyvers's algorithm
\cite{griffiths2004finding} operates very differently. Rather than estimating
$\theta$ and $\phi$, it learns instead the posterior distribution over the
assignments of words to topics $P(z\mid w)$, and it does so with the help of
Gibbs sampling, thus assigning topics to each word. After a certain number of
sampling iterations (the "burn-in"), these assignments are a good indicator of
there being a relationship between word and topic, and between topic and
document. From them, we can pick the topics have been assigned to a given word
in its various instanciations, and retrieve the documents to which these topics
have been assigned. Furthermore, when necessary, $\phi$ and $\theta$ can be
calculated from the assignments.

\subsection{Concept Presence in Topics}

We assume that the presence of a concept in a topic is indicated by the presence of a
word typically associated with the concept in question. Therefore a topic’s
association with a word is indicative of its association with the corresponding
concept. The LDA model explicitly links words to topics, but in a graded way:
each word is associated with each topic to a certain degree. From this
information, we can use various heuristics to rule whether a concept is involved
in a topic or not.

In this study, we tested these heuristics:
\begin{description}
    \item[Most Likely:] The queried concept is associated to the topic which
    makes its corresponding word most likely to occur.

    \item[Highest Rank:] The queried concept is associated with the topic in
    which its corresponding word has the highest rank on the topic's list of
    most likely words.

    \item[Top 30 Rank:] The queried concept is associated with the topics in
    which its corresponding word is among the top 30 words on the topic's list
    of most likely words.

    \item[Concrete Assignment:] In the Gibbs Sampling method, individual words
    are assigned to topics, and word likelihood given a topic is calculated from
    these assignments. We can thus say that a word is involved in a topic if
    there is at least one assignment of this topic to this word in the corpus.
\end{description}

Using these heuristics and an LDA model (learned using either Hoffman's or
Griffiths \& Steyvers's method), we can determine for a given concept the topics
in which it is involved.

Depending on the learning method, we can then determine which textual segments
are associated to a given topic. On one hand, in Hoffman's method, when a topic
is assigned to a segment, there will be a non-zero probability that any given
word in the segment is associated with the topic in question. On the other hand,
when learning the LDA model using Gibbs Sampling, we'll consider that a topic is
associated to a textual segment if there is at least one word of this segment
that is associated with the topic in question.

Thus, from a given concept, we can retrieve the segments in which the concept is likely
expressed by retrieving the textual segments that are associated to the topics
which are associated to the queried concept.

\section{Experimentation\label{sec:Exp}}

\subsection{Corpus\label{subsec:Corpus}}

Algorithms were tested on a French-language corpus of 5,229 decisions from the
{\em Cour d'appel du Québec} (Quebec Court of Appeal), the highest judicial
court in Quebec. Much like philosophical discussions, arguments in juridical
texts, and in decisions in particular, are well-developed, and nuances are
important, so we can expect concepts to be explained thoroughly and employed
with precision. However, there is much more homogeneity in style and vocabulary,
and this style and vocabulary are more familiar to the broader public than in
typical philosophical works, which facilitates annotation. Thus, court decisions
are likely to afford complex conceptual analyses, but lack the difficulties that
come with the idiosyncrasies of individual philosophical texts.

Court decisions were divided into paragraphs, yielding 198,675 textual segments, which
were then broken down into words and lemmatized using TreeTagger
\cite{schmid1994probabilistic}. Only verbs, adjectives, nouns and adverbs were
kept, and stopwords were removed.

In order to provide a gold standard against which we could evaluate
the performances of the chosen algorithms, annotations were collected
using CrowdFlower\footnote{\url{http://www.crowdflower.com}}.

In a first "tagging" step, French-speaking participants were given
a textual segment and were instructed to write down five concepts which are
expressed in the segment\textemdash more specifically, the criterion mentioned
in the instructions was that the concept must contribute to the discourse
(in French: "\emph{propos}") expressed in the segment. 25 participants annotated 105 segments
in this way, yielding 405 segment annotations for a total of 3,240
segment-concept associations.

Data obtained from this first step can tell us that a concept is present
in a segment, but we can never infer its absence from it, as its absence
from the annotations could simply mean that the annotator chose to
write down five other concepts and had no more place for another one.
Therefore, it was necessary to add another step to assess absence.

In the second "rating" step, participants were given a segment and six
concepts (from the pool of concepts produced in the tagging step), and were
instructed to rate each concept's degree of presence or absence from 1 (absent)
to 4 (present). The degree of presence is meant to give options to the
participant to mark a concept as present, but to a lesser degree, if, say, it
is not particularly salient, or if lack of context gives way to some doubt as to
whether it really is present. Using this strategy, we can get participants to
mark the absence of a concept (degree 1 of the scale) in a way that is intuitive
even if one has not properly understood the instructions. For our purposes, we
assume that CrowdFlower participants mark a concept as absent when they give it
a rating of 1, and as present (even if minimally) if they make any other choice.
After removing low-quality annotations, we get 104 segments annotated by 37
participants, for a total of 5,256.

\begin{table}[t]
\begin{centering}
\begin{tabular}{cccc}
 &  & \multicolumn{2}{c}{\textbf{Legal experts}}\tabularnewline
 &  & \textbf{Present} & \textbf{Absent}\tabularnewline
\hline
\multirow{2}{*}{\textbf{CrowdFlower participants}} & \textbf{Present} & 32 & 2\tabularnewline
 & \textbf{Absent} & 24 & 4\tabularnewline
\end{tabular}
\par\end{centering}
\caption{Contingency table of the CrowdFlower ratings against the legal experts'
ratings for the rating step. \label{Confusion-matrix-of}}
\end{table}

In order to ensure that annotations by CrowdFlower participants reflect a
genuine understanding of the text, we also recruited legal experts to make
similar taggings and judgments and to compare annotations. While the first task was
the same for the experts, the second was slightly different in that there were
only two options, and in that they were given oral and written instructions to only
mark as absent concepts which were definitely absent. This is because the contact
we had with these participants made it possible to ensure that instructions were
well understood: we did not need to add options to reinforce the idea that a
concept is only absent when it is completely and undoubtedly absent. In total, 5
experts tagged 82 text segments in the tagging step, producing a total of
361 tag-segment pairs, and 4 experts rated concepts on 58 segments in the
rating step, producing a total of 412 tag-segment pairs.

As Table \ref{Confusion-matrix-of} shows, the distribution is skewed towards
presence, which makes Cohen's $\kappa$ a poor choice of metric
\cite{gwet2008computing}. Gwet's AC1 coefficient \cite{gwet2008computing} was
used instead, and it revealed that CrowdFlower participants and legal experts
have moderate but above-chance agreement, with a coefficient of 0.30 and
$p$-value of less than 0.05 (indicating that there is less than 5\% chance that
this above-chance agreement is due to random factors)%
\footnote{The scenario on the tagging step does not fit any of the common
inter-annotator agreement metrics. Firstly, a single item is given five values
for the same property. Secondly, in our annotations, absence of annotation does
not mean absence of concept; the converse would have been a common assumption in
inter-annotator metrics.}.
As the confusion matrix of Table
\ref{Confusion-matrix-of} shows, the error mostly comes from the fact that
CrowdFlower participants seem much more likely to mark concepts as absent than
legal experts.

\subsection{Algorithms\label{subsec:Algorithms}}

Both LDA algorithms were implemented as described in the previous section. For
the online LDA, we have used the implementation that is part of Gensim
\cite{884893}, and for the Gibbs sampler-LDA, we have adapted and optimized code
from Mathieu Blondel \cite{blondel_latent_2010}. In both cases, we used $k=150$
topics as parameter, because observing the semantic coherence of the most
probable words in each topic (as indicated by $\phi_{z}$) suggests that
greater values for $k$ yield topics that seem less coherent and less
interpretable overall. For the Gibbs sampler-LDA, we did a burn-in of 150
iterations.

The baseline chosen was the keyword heuristic: a concept is marked
as present in a segment if the segment contains the word that represents
it, and absent if it does not.

Each method was successively applied to our corpus, using, as queries, items from
a set of concept-representing words that were both used in annotations from the
rating steps and found in the corpus lexicon. In total, this set numbers 229
concepts for the legal experts' annotations and 808 concepts for the CrowdFlower
annotations. Among these, 170 terms are found in both sets of annotations.

\section{Results\label{sec:Results}}

\begin{table}[t]
\begin{centering}
\begin{tabular}{cccccccc}
 &  & \multicolumn{3}{c}{\textbf{CrowdFlower}} & \multicolumn{3}{c}{\textbf{Experts}}\tabularnewline
 &  & \textbf{Recall} & \textbf{Precision} & \textbf{F1} & \textbf{Recall} & \textbf{Precision} & \textbf{F1}\tabularnewline
\hline
 & \textbf{Keyword} & 0.03 & 0.56 & 0.07 & 0.01 & \textbf{1.00} & 0.04\tabularnewline
\hline
\multirow{3}{*}{\textbf{Online LDA}} & \textbf{Most Likely} & 0.06 & 0.63 & 0.13 & 0.03 & 0.67 & 0.07\tabularnewline
 & \textbf{Highest Rank} & 0.07 & 0.51 & 0.16 & 0.03 & 0.50 & 0.07\tabularnewline
 & \textbf{Top 30 Rank} & \textbf{0.18} & 0.60 & \textbf{0.32} & \textbf{0.15} & 0.61 & \textbf{0.29}\tabularnewline
\hline
\multirow{4}{*}{\textbf{\vbox{\hbox{\strut Gibbs}\hbox{\strut Sampling-}\hbox{\strut LDA}}}} & \textbf{Most Likely} & 0.05 & 0.55 & 0.12 & 0.05 & 0.50 & 0.13\tabularnewline
 & \textbf{Highest Rank} & 0.00 & 0.60 & 0.01 & 0.03 & 0.50 & 0.07\tabularnewline
 & \textbf{Top 30 Rank} & 0.01 & 0.64 & 0.03 & 0.01 & 0.25 & 0.04\tabularnewline
 & \textbf{Concrete Assign} & 0.08 & \textbf{0.65} & 0.19 & 0.12 & 0.53 & 0.25\tabularnewline
\hline
\end{tabular}
\par\end{centering}
\caption{Performance for each method, calculated using data from the rating
task. \label{F1-step2}}
\end{table}

Results from the application of the baseline and our methods on all concepts
were compared to the gold standards obtained from the rating step using overall
precision, recall, and F1-score. They are illustrated in Table \ref{F1-step2}.

Apart from the Gibbs Sampling-LDA/Highest Rank method, all of the proposed methods
improved on the baseline, except for the ones using word rankings among the
Gibbs Sampling-LDA methods. This is due in particular to improvements in recall.
This is to be expected, as the keyword only targets one way in which a concept
can be expressed, and thus appears to be overly conservative.

Among the Gibbs Sampling-LDA methods, Concrete Assignment fares significantly
better, but the best overall, both in recall and F1-score, is the Online LDA/Top
30 Rank. On this, experts and non-experts are in agreement.

\section{Discussion\label{sec:Discussion}}
These results seem to validate this study's main hypothesis, that is, LDA
methods can improve on the keyword heuristic when it comes to detection of
concept expression.

This said, recall remains under 20 \%, indicating that topic models are still
insufficient to detect all forms of expression of a concept. As such, while it
is a clear improvement on the keyword heuristic, it would seem to contradict our
hypothesis that topic models can be used to detect all sorts of concept
expressions.

\begin{table}[t]
\begin{centering}
\begin{tabular}{ccc}
 & \textbf{CrowdFlower} & \textbf{Legal experts}\tabularnewline
\hline
\textbf{Tagging task (step 1)} & 0.35 & 0.10\tabularnewline
\textbf{Rating Task (step 2)} & 0.75 & 0.24\tabularnewline
\end{tabular}
\par\end{centering}
\caption{Reuse rate in annotation tasks. \label{Reuse-rate}}

\end{table}

\subsection{Quality of Annotations}

While experts' and non-experts' annotations are mostly in agreement, there are
important discrepancies. Experts' annotations systematically give better scores
to Gibbs Sampling methods, and lower scores to Online LDA methods, than
non-experts'. For instance, while the Online LDA/Top 30 Rank method beats the
Gibbs Sampling/Assignment method by 0.11 in F1-scores using CrowdFlower
annotations, this difference shrinks to 0.4 when using experts' annotations.
These discrepancies, however, can be traced to a difference in types of
heuristics employed in the tagging step: CrowdFlower participants are more
likely to employ words from the excerpt as annotations (i.e. using the concept
BROTHER when the word ``brother'' is present {\em verbatim} in the text
segment), which favors the baseline.

In order to give evidence for this claim, we calculated the propensity of a
participant to mark as present a tag that is also a word in the text segment. Specifically, we estimated
the reuse rate\footnote{The reuse rate here is simply the number of tags which
are a word in the text segment divided by the total number of tags that are words.
Multi-word expressions were excluded because detecting whether they are in the
text or not would be complicated.} as depicted by Table \ref{Reuse-rate}.\footnote{%
Experts’ annotations were ignored because there were too few annotation instances
where the queried concept's keyword was in the textual segment, and, as a
result, values for the "Keyword in segment" condition were uninformative.
}).
As it turns out, in
the initial tagging step, CrowdFlower participants are more than three times
more likely to write down a word that is present in the text. As participants
in the rating step are only rating tags entered by people of the same group,
this translates into a similar ratio in the rating task. However, as it seems
that in the rating step, participants are less likely to mark as present a word
which is not specifically in the text, reuse rate is inflated for both
participant groups. As a result, a large majority of one-word annotations by
CrowdFlower participants are already in the text, while the reverse is still
true of expert annotations.

Thus, when we discriminate between tags that are present in the textual segment
and those that are not, we get a much clearer picture (Table \ref{F1-no-cooc}). In
the first case, the best heuristic is still the baseline, with Online LDA
methods offering much better results than Gibbs Sampling-LDA methods. But in the
second, the baseline is unusable, and while F1-scores of Online LDA methods drop
by more than half, Gibbs Sampling-LDA methods stay the same or improve. Having fewer
annotations where the concept's keyword is in the textual segment will penalize
the Online LDA methods, but not the Gibbs Sampling-LDA ones.

As such, this discrepancy should not count as evidence against the hypothesis
that CrowdFlower annotations are invalidated by their discrepancies with
experts' annotations. However, it suggests that future annotations should control
for the ratio of present and absent words in the rating step. Furthermore, it
would be useful to test participants of a same group on the same textual
segment/concept pair in order to compare in-group inter-annotator agreement with
between-group inter-annotator agreement.

\begin{table}[t] \begin{centering} \begin{tabular}{cccc} &  & \textbf{Keyword in
segment} & \textbf{Keyword absent}\tabularnewline \hline & \textbf{Keyword} &
\textbf{0.72} & 0.00\tabularnewline \hline \multirow{3}{*}{\textbf{Online LDA}} &
\textbf{Most Likely} & 0.21 & 0.13\tabularnewline & \textbf{Highest Rank} & 0.48 &
0.14\tabularnewline & \textbf{Top 30 Rank} & 0.67 &
\textbf{0.30}\tabularnewline \hline \multirow{4}{*}{\textbf{\vbox{\hbox{\strut
Gibbs}\hbox{\strut Sampling-}\hbox{\strut LDA}}}} & \textbf{Most Likely} & 0.00 &
0.12\tabularnewline & \textbf{Highest Rank} & 0.00 & 0.01\tabularnewline &
\textbf{Top 30 Rank} & 0.00 & 0.03\tabularnewline & \textbf{Concrete
Assignment} & 0.21 & 0.19\tabularnewline \hline \end{tabular}
\par\end{centering} \caption{F1-scores against CrowdFlower annotations for each
method, based on presence or absence of the queried concept keyword in the
textual segment. \label{F1-no-cooc}}

\end{table}

\subsection{Improving on Topic Model Methods}

In any case, while it does not solve the problem of retrieving all the textual
segments where a concept is expressed, the Online LDA/Top 30 Rank method makes
important headway towards a more satisfactory solution. It improves on the
keyword heuristic's F1-score by 0.25 (both when experts' and non-experts'
annotations are used as gold standard), and, as such, constitutes a clear
improvement and a much better indicator of concept presence.

Improvements could be reached by associating different approaches to concept
detection, when we know that some methods do better than others in specific
contexts. For example, the keyword heuristic does slightly better than Online
LDA/Top 30 Rank when the queried concept's keyword is present in the text, so
it could be used in these situations, while the former method could be used in
other cases. In fact, this produces a minor improvement (F1-score of 0.33 with
the CrowdFlower gold standard, as compared to 0.32 for pure Online LDA/Top 30
Ranks). We can hope that including other methods for other means of expressing a
concept can contribute to further improvements.

\section{Conclusion\label{sec:Concl}}

In this paper, we expressed the problem of concept detection for the purpose of
philosophical conceptual analysis, and sought LDA-based methods to address it.
In order to evaluate them, we devised an annotation protocol and had experts and
non-experts annotate a corpus.

Our results suggest that LDA-based methods and the Online LDA/Top 30 Rank method
in particular, can yield important improvements over the keyword heuristic that
is currently used as a concept detection heuristic in many contexts. Despite
important improvement, it remains a high-precision, low-recall method. However,
while more work remains to be done, this indicates that detecting concepts
through topics can serve as a general-purpose method for at least some forms of
concept expression that are not captured using naive keyword approaches.

As such, we suggest that further research should try to integrate other methods
of detecting concept presence in textual data that focus on other means of
expressing concepts in texts and discourse.

\subsubsection*{Acknowledgments.} This work is supported by research grants from
the Natural Sciences and Engineering Research Council of Canada (NSERC) and from
the Social Sciences and Humanities Research Council of Canada (SSHRC).

\end{document}